\documentclass[conference]{IEEEtran}
\IEEEoverridecommandlockouts
\usepackage{cite}
\usepackage{amsmath,amssymb,amsfonts}
\usepackage{algorithmic}
\usepackage{graphicx}
\usepackage{textcomp}
\usepackage{xcolor}

\usepackage{subcaption}
\usepackage{comment}
\usepackage{tabularx}

\usepackage{cite}
\usepackage{amsmath,amssymb,amsfonts}
\usepackage{algorithmic}
\usepackage{textcomp}
\usepackage{xcolor}

\usepackage{amsthm,enumitem}

\usepackage{soul}
\usepackage{multirow}

\def\BibTeX{{\rm B\kern-.05em{\sc i\kern-.025em b}\kern-.08em
    T\kern-.1667em\lower.7ex\hbox{E}\kern-.125emX}}
\begin{document}

\title{Impact of Image Context for Single Deep Learning Face Morphing Attack Detection\\
}

\author{\IEEEauthorblockN{Joana Alves Pimenta}
\IEEEauthorblockA{\textit{Institute of Systems and Robotics} \\
\textit{University of Coimbra}\\
Coimbra, Portugal \\
joana.pimenta@isr.uc.pt}
\and
\IEEEauthorblockN{Iurii Medvedev}
\IEEEauthorblockA{\IEEEauthorrefmark{1}\textit{Institute of Systems and Robotics} \\
\textit{University of Coimbra}\\
Coimbra, Portugal \\
iurii.medvedev@isr.uc.pt}
\and
\IEEEauthorblockN{Nuno Gonçalves \IEEEauthorrefmark{1}\IEEEauthorrefmark{2}}
\IEEEauthorblockA{\IEEEauthorrefmark{2}\textit{Portuguese Mint and Official , } \\
\textit{Printing Office (INCM)}\\
Lisbon, Portugal \\
nunogon@deec.uc.pt}
}

\maketitle

\begin{abstract}
The increase in security concerns due to technological advancements has led to the popularity of biometric approaches that utilize physiological or behavioral characteristics for enhanced recognition. Face recognition systems (FRSs) have become prevalent, but they are still vulnerable to image manipulation techniques such as face morphing attacks. 
This study investigates the impact of the alignment settings of input images on deep learning face morphing detection performance. We analyze the interconnections between the face contour and image context and suggest optimal alignment conditions for face morphing detection.
\end{abstract}


\section{Introduction}
\vspace{0cm}
The expansion of technological advancements in modern society has led to an increase in security concerns. Traditional identification methods have become less reliable due to their vulnerability to forgetfulness, loss, replication, or theft, thereby compromising their intended security function. As a solution to this issue, biometric approaches are gaining popularity as they utilize physiological or behavioral characteristics to enhance the recognition process. 
Face image modality took one of the most important roles in modern biometric applications due to the simplicity of face image acquisition and recent advances in computer vision techniques. This led to the widespread use of Face Recognition Systems (FRSs) which utilize facial traits for the purpose of identification or verification \cite{9145558}. Despite the fact that FRSs are currently used in various applications, they are still highly vulnerable to attacks due to the extensive range of image manipulation techniques that can be used to deceive the system. 

One of the most important types of threats to FRSs is the face morphing attack. In this attack, facial features from two or more images are merged to create a synthetic image that incorporates features from both faces. The resulting image is similar to the images that gave rise to it, which allows one person to impersonate another, thereby violating the principle of self-ownership.
That is why face morphing detection is a critical task in the era of digital manipulation and deep learning techniques. 
However, the performance of face morphing detection may depend on various factors, such as the alignment and preprocessing of input images. 
Specifically, the face image alignment setting can impact the amount of context included in the input image, which in turn can hypothetically affect the performance of the detection algorithm. We conduct our research to define optimal alignment settings for face morphing detection, exploring the possibility of using interconnections between the face contour and image context to improve the performance of the detection algorithm.

Essentially, our purpose is to investigate the relationship between image context and MAD, with the aim of identifying the most effective context properties for detection. Throughout this paper, the term ”image context” refers to the background and surrounding elements in the image, i.e., the part of the image that does not contain the face.

As an additional contribution, we combined a dataset that adheres to the International Civil Aviation Organization (ICAO) guidelines for detecting face morphing. 

\vspace{0cm}
\section{Related Work}

\textbf{Face Recognition.} Current advances in face recognition methods use deep learning techniques that employ deep neural networks, allowing the learning of deep facial features, which have high discriminative power. 


Face recognition deep networks are commonly trained using classification-based tasks, employing softmax loss or its margin-based alternatives like ArcFace \cite{8953658}. The addition of a margin to the softmax loss is crucial because it significantly improves the discriminative power of the learned features.
More recently, there has been a focus on incorporating adaptiveness into the margin based on the quality of the input image. For instance, MagFace \cite{meng2021magface} optimizes the feature embedding using an adaptive margin and regularization based on its magnitude. Another approach is AdaFace \cite{kim2022adaface}, which proposes adapting the margin function based on the norm of the feature embedding.


\textbf{Face Morphing Generation.}
Face morphing can be performed using landmark-based or deep learning-based approaches. 
Landmark-based methods employ a set of fiducial facial points, which are detected on all contributing face images, to generate a morph image by warping and bending procedures \cite{ferrara2014magic}. 

Deep learning-based methods may employ encoder-decoder architectures, such as Generative Adversarial Networks (GANs) \cite{NIPS2014_5ca3e9b1}.
For example, the MorGAN \cite{8698563} approach aims to make the generated images look similar to the real images while also encouraging the generators to produce diverse images that differ from each other.
Karras et al. \cite{StyleGAN} proposed the StyleGAN approach, which can be used to generate high-quality morphs. 


The MIPGAN \cite{mipgan} approach revisits the StyleGAN by introducing an end-to-end optimization approach with a novel loss function that emphasizes preserving the identity of the generated morphed face images by incorporating identity priors.
MorDIFF \cite{MorDIFF} proposes the use of diffusion autoencoders to generate high-fidelity and smooth face morphing attacks, which are highly vulnerable to state-of-the-art face recognition models. 
ReGenMorph \cite{ReGenMorph} approach proposes to eliminate blending artifacts by combining image-level morphing and GAN-based generation, resulting in visibly realistic morphed images with high appearance quality.


\textbf{Face Morphing Detection.}
Morphing attack detection (MAD) methods can be classified into two types, depending on the security application scenario: Single Morphing Attack Detection (S-MAD) and Differential Morphing Attack Detection (D-MAD). 


S-MAD refers to techniques that can detect a morphed image without comparing it to an authentic reference image (\textit{non-reference}). They are therefore based on the analysis of visual artifacts or inconsistencies in the morphed image itself.
Many approaches rely on the analysis of handcrafted features like Binarized Statistical image features (BSIF) \cite{7791169}, Local Binary Pattern (LBP) \cite{OJALA199651}, Local Phase Quantization (LPQ) \cite{Blur_Texture_Classification} image descriptors, and Photo Response Non-Uniformity (PRNU) known as sensor noise \cite{Morphing_detection_PRNU}. 

Recent works intensively uses deep learning for face morphing detection. 
OrthoMAD approach \cite{orthomad} proposes to use a regularization term for the creation of two orthogonal latent vectors that disentangle identity information from morphing attacks. 
MorDeephy method \cite{MorDeephy} introduced fused classification to generalize morphing detection to unseen attacks. The formulation will be followed in this work.
Tapia et al. \cite{Tapia2021SingleMA} proposed a framework using few-shot learning with siamese networks and domain generalization. The framework includes a triplet-semi-hard loss function and clustering to assign classes to image samples. 
In this work, we focus only on the S-MAD case to perform the analysis of image alignment settings.

\vspace{0mm}
\section{Methodology}

\vspace{0mm}
\textbf{Source Data Curating}. 
An initial challenge encountered in this research was the lack of a suitably extensive dataset that conformed to ICAO compliance requirements. To address this issue, we combined multiple datasets comprising compliant images, including both publicly available and privately obtained data. When selecting the datasets, we prioritized those that provided a larger number of images per identity and included the following ones: FRGC \cite{51406}, XM2VTS \cite{xm2vtdb}, ND Twins \cite{50686}, FERET \cite{879790,PHILLIPS1998295}, AR \cite{ardata}, PICS \cite{pics}, FEI \cite{feidata}, IMMF \cite{IMM2005-03943} and GTDB \cite{Ggtdb}.

Several selected components were filtered to remove non-compliant images, i.e., non-frontal images or other images not suitable for morphing.
In the specific case of the ND twins dataset, only one twin from the pair was included due to their striking resemblance, which will be confusing for the methodology of this research.
Our result dataset, which we call the ICMD dataset, comprises over 50k images of more than 2.5k individuals. 

\textbf{Morph Image Generation}. 
To accompany our training data with face morph samples, we employed landmark-based and deep learning-based (specifically GAN-based) face morphing approaches. These samples are generated using the originals from the ICMD dataset, where pairing is performed following the \cite{MorDeephy}, to ensure  unambiguous class labeling in the fused classification task.

Namely, the identity list of the dataset is randomly split into two disjoint subsets attributed to the First and Second networks, and the pairing is made between those subsets. In the end, we ensured a consistent classification classification of morphed combinations by the networks. 

To generalize the detection performance and reduce overfitting for artifact detection, we have included \textit{selfmorphs} for both LDM and StyleGAN approaches.  \textit{Selfmorphs} are generated using images of the same individual, resulting in morphed images that continue to represent that same individual but contain merging artifacts of a different kind. 
As a result, considering \textit{selfmorphs} as \textit{bona fide} samples we can prioritize morphing detection based on the behavior of deep facial features.


\textbf{Alignment settings} 
Our search for the optimal amount of image context for morphing detection is based on selecting several different alignment settings and running identical experiments for each setting. The face alignment in academia is usually performed by a rigid transformation, which minimizes the coordinate distance between the five facial landmarks (detected by MTCNN\cite{mtcnn}) (\{left eye\}, \{right eye\}, \{nose\}, \{left mouth corner\}, \{right mouth corner\}) and the definite target list of coordinates (for the resulting image size of $112\times112$ - \{\{38.2, 41.7\}, \{73.5, 41.5\}, \{56.0, 61.7\}, \{41.5, 82.4\}, \{70.7, 82.2\}\}) \cite{8953658}. The particular list of settings that we used is based on the scaling of this target set of coordinates.
The Table \ref{Tab1} presents a schematic correspondence of each alignment with the scale factor utilized, along with its respective indicative ratio of the face's occupancy area in the image. We estimate this face's occupancy as the ratio of face area (limited by a face contour detected using 68 landmarks \cite{dlib09}) to the full image area.
\vspace{0mm}
\begin{table*}[h]
\centering
\caption{Summary table of all alignment conditions with their respective scale factors and ratios.}
\begin{tabular}{|c|c|c|c|c|c|c|c|c|c|c|c|}
\hline
\textbf{Alignments} &
  \textbf{a} &
  \textbf{b} &
  \textbf{c} &
  \textbf{d} &
  \textbf{e} &
  \textbf{f} &
  \textbf{g} &
  \textbf{h} &
  \textbf{i} &
  \textbf{j} &
  \textbf{k} \\ \hline
\textbf{Scale Factor} &
  1.65 &
  1.40 &
  1.10 &
  1.00 &
  0.90 &
  0.85 &
  0.80 &
  0.75 &
  0.70 &
  0.65 &
  0.60 \\ \hline
\textbf{Ratio} &
  0.15 &
  0.21 &
  0.34 &
  0.42 &
  0.51 &
  0.56 &
  0.62 &
  0.70 &
  0.77 &
  0.86 &
  0.94 \\ \hline
\end{tabular}%
\label{Tab1}
\end{table*}

\vspace{0mm}

\begin{figure*}[!ht]
\centering
    \includegraphics[width=0.75\linewidth]{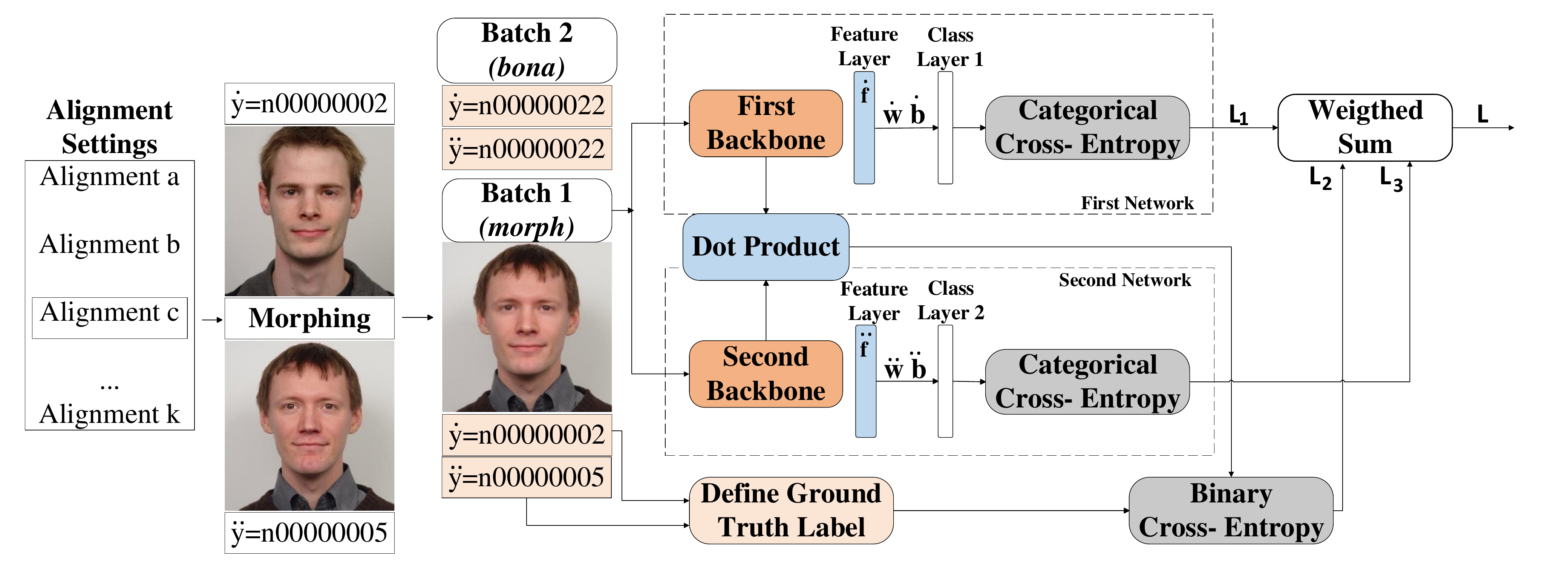}
    \caption{\label{logo} S-MAD fused approach schematics. In order to simplify the visualization, a single image is shown per batch.}
\end{figure*}

\textbf{S-MAD - Fused Classification.} In our work, we approach \textit{no-reference} face morphing detection in several ways. First, we follow the fused classification approach, where two parallel networks were trained simultaneously. These networks were specifically designed to acquire high-level features by performing classification tasks in order to generalize the performance to unseen attacks \cite{MorDeephy}.

%
%
%
%

The overall pipeline schematic is presented in Fig.\ref{logo}. 
Each sample is assigned two class labels: \textit{morphs} inherit them from source identities; \textit{bona fides} have a duplicated original label. The classification task is made differently for each of the networks. First Network labels them by the original identity from the first source image, and the Second Network by the second original label. The main motivation is learning high-level identity discriminative features, which can indicate the presence of face morphing. Such classification is regulated by the explicit binary classification of a dot product of those resulting high-level features.


Mathematically, such a schematic is formulated as the weighted sum: $L = \alpha_{1} L_{1} + \alpha_{2} L_{2} + \beta L_{3}$, where $L_{1}$ and $L_{2}$ are face recognition components, and $L_{3}$ is a morphing detection component.  Based on the common softmax formulation, each network is regularized by the respective losses:
\vspace{0mm}
\begin{equation}
\label{equation:loss1}
    L_{1} = - \frac{1}{N}\sum_{i}^{N} \log (\frac{e^{\dot{W}_{\dot{y}_i}^{T}\dot{f}_{i}+\dot{b}_{\dot{y}_i}}}{ \sum_{j}^{C} e^{\dot{f}_{\dot{y}_j}}})
\end{equation}
\begin{equation}
\label{equation:loss2}
    L_{2} = -\frac{1}{N}\sum_{i}^{N} \log (\frac{e^{\ddot{W}_{\ddot{y}_i}^{T}\ddot{f}_{i}+\ddot{b}_{\ddot{y}_i}}}{ \sum_{j}^{C} e^{\ddot{f}_{\ddot{y}_j}}}),
\end{equation}
\vspace{0mm}
where $f_i$ are deep features of the $i^{th}$ sample, $y_i$ represents the class index of the $i^{th}$ sample, and $W$ and $b$ denote the weights and biases of the last fully connected layer, respectively. $N$ represents the batch size, while $C$ represents the total number of classes.

Finally, in order to determine the similarity metric based on the ground truth authenticity label of the image, the morphing detection score is obtained by computing the dot product of the backbone outputs ($\dot{f} \cdot \ddot{f}$). This score is then passed through the \textit{sigmoid} function and used to define the binary cross-entropy loss.
As a final result, the corresponding loss is defined by:
\vspace{0mm}
\begin{equation}
\label{equation:loss3}
    L_{3} = -\frac{1}{N} \sum_{i}^{N} t\log \frac{1}{1+e^{-\ddot{f} \cdot \dot{f}}}
    + (1-t)\log \left( 1-\frac{1}{1+e^{-\ddot{f} \cdot \dot{f}}} \right)
\end{equation}
\vspace{0mm}
The optimization process involves combining the resulting losses as a weighted sum, resulting in $L$, with the goal of minimizing it. This is done to learn facial features that are discriminative and specifically regularized for the task of detecting morphing.

\begin{figure*}[ht!]
\centering
    \includegraphics[width=0.75\linewidth]{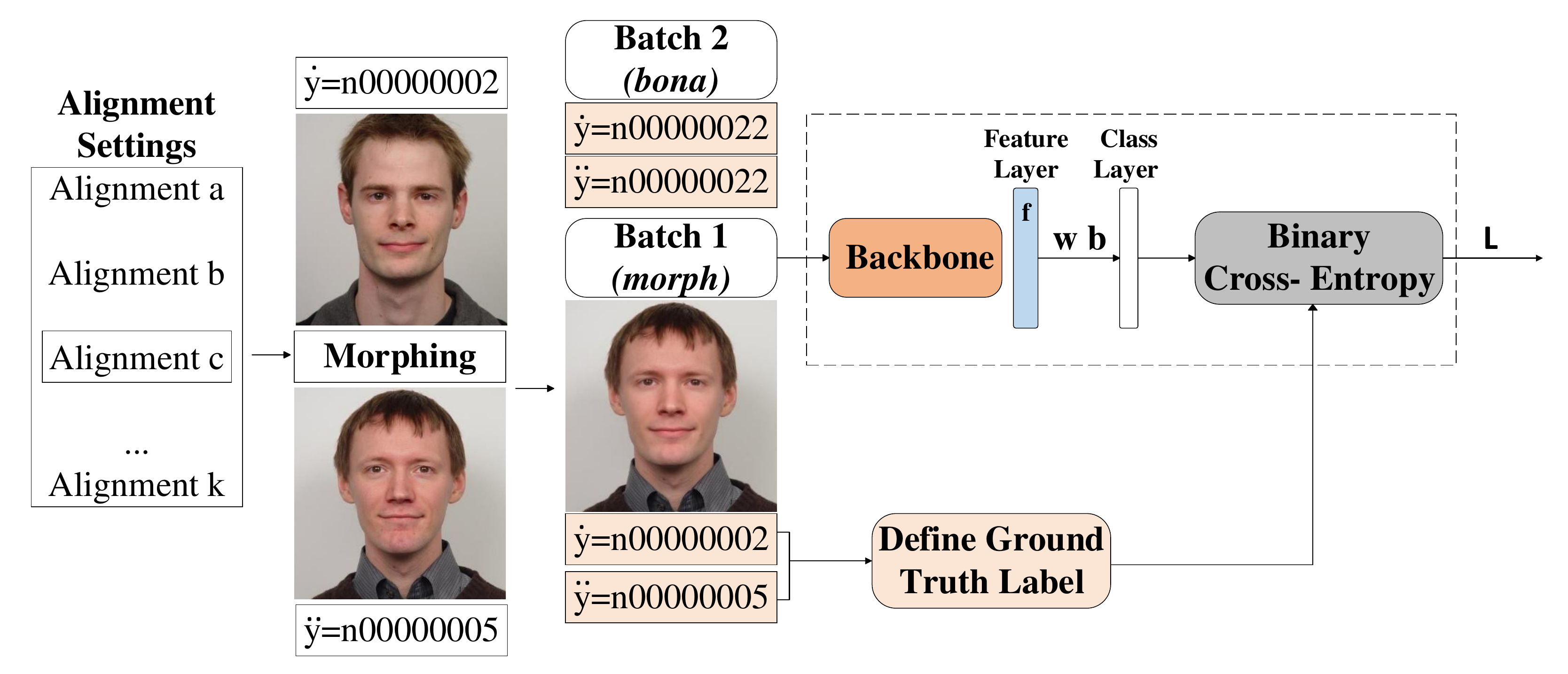}
    \caption{\label{logo1} S-MAD approach model schema for a single network. In order to simplify the visualization, a single image is shown per batch.}
\end{figure*}

\textbf{S-MAD - Binary Classification.} Another approach for face morphing detection is indeed similar to the straightforward binary classification (morph/non-morph). To implement it, we removed the identity classification part from the fused approach and retained only a single deep network in the entire pipeline. The model schema is presented in Fig.\ref{logo1}.

\textbf{Benchmarking.} For performance estimation, we employ the open-source morphing benchmarking utilities \footnote{https://github.com/iurii-m/MorDeephy.git} and adopt them into our work. We replace the \textit{bona fide} subset with the images from  FRLL-Set\cite{FRLL}, Utrecht\cite{pics}, MIT-CBCL\cite{MIT-CBCL} and EFIEP\cite{EFIEP} (since the default suggested protocols share images with our training data).
All protocols share the same list of \textit{bona fide} images and are only different in the content of morphs, which are taken from the FRLL-Morphs dataset \cite{FRLL-Morphs} (protocol names correspond to the FRLL-Morph subset names):
\textit{protocol-asml} with $\sim$ 2k morphs, \textit{protocol-opencv} with $\sim$ 1.3k morphs, \textit{protocol-facemorpher} with $\sim$ 2k morphs, \textit{protocol-webmorph} with $\sim$ 1k morphs and \textit{protocol-stylegan} with $\sim$ 2k morphs.

    

    



\textbf{Heatmap Computation.} We analyze the image context impact using the Gradient-Weighted Class Activation Mapping (Grad-CAM) technique and generate a heatmap that highlights the regions of the input image that have the most significant influence on the ground truth binary prediction.

\vspace{0cm}
\section{Experiments and Results}

\vspace{0mm}
\textbf{Training Settings.} As a baseline model in our work, we use EfficientNetB3  \cite{efiicient}, which is pretrained on the ImageNet dataset. 
We trained our models for five epochs using a stochastic gradient descent (SGD) optimizer with a momentum of 0.9 and a learning rate linearly decaying from 0.075 to 1e-5 . The batch included 28 images. 
Separate training experiments are performed for each alignment case on concatenated datasets: original, LDM, StyleGAN morphs, and \textit{selfmorphs}. Face morphs are generated with LDM and StyleGAN approaches.
The parameters for the fused approach, which determine the appropriate balance between the different components of the loss function, are taken from the original work \cite{MorDeephy}: $\alpha$= $\alpha_{1}$= $\alpha_{2}$ and $\alpha / \beta$=0.2. 

\begin{figure}[!ht]
\centering
    \includegraphics[width=0.98\linewidth]{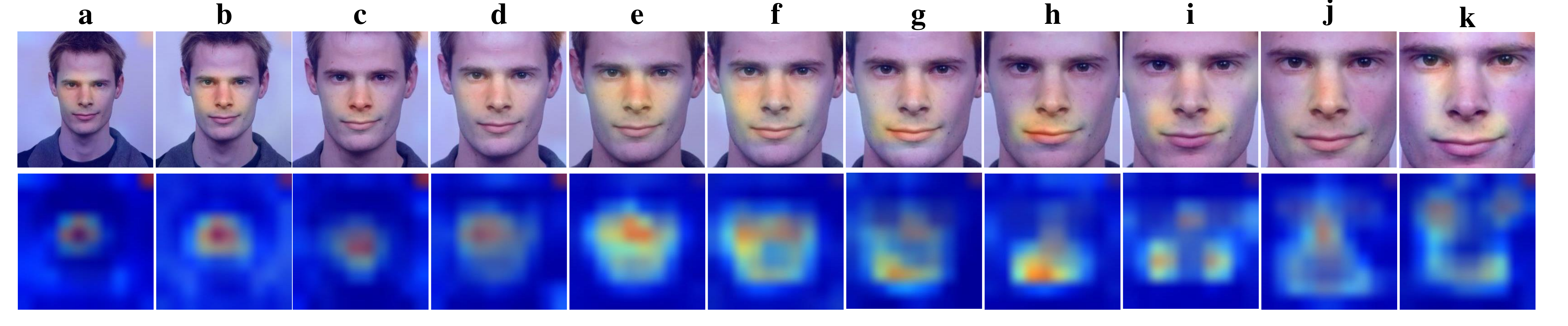}
    \caption{\label{gradbin} Grad-CAM morph heatmaps for the S-MAD binary approach under different alignment conditions (Recall that \textit{bona fide} sets are equal across all the protocols).
}
\end{figure}

\textbf{Binary Classification.} 
Based on the results presented in Table \ref{allmodelBIN}, the alignment range with optimal performance is observed between \textit{e} to \textit{g}, with \textit{e} being the possible optimal case. 
Based on heatmaps, the face is the principal region for the detection decision, and the regions, which are prompt to contain morphing artifacts, are mainly activated (see Fig. \ref{gradbin}). 

\begin{table*}[ht!]
\centering
 \caption{BPCER@APCER = (0.1, 0.01) of our S-MAD binary approach for various alignment settings}
\label{tab:my-table}
\begin{tabular}{|c|cccccccccc|ll}
\cline{1-11}
\multirow{3}{*}{\textbf{Alignments}} &
  \multicolumn{10}{c|}{\textbf{BPCER@APCER=$\delta$}} &
   &
   \\ \cline{2-11}
 &
  \multicolumn{2}{c|}{\textbf{Protocol-asml}} &
  \multicolumn{2}{c|}{\textbf{Protocol-facemorpher}} &
  \multicolumn{2}{c|}{\textbf{Protocol-opencv}} &
  \multicolumn{2}{c|}{\textbf{Protocol-stylegan}} &
  \multicolumn{2}{c|}{\textbf{Protocol-webmorph}} &
   &
   \\ \cline{2-11}
 &
  \multicolumn{1}{c|}{\textbf{$\delta$=0.1}}  &
  \multicolumn{1}{c|}{\textbf{$\delta$=0.01}}  &
  \multicolumn{1}{c|}{\textbf{$\delta$=0.1}}  &
  \multicolumn{1}{c|}{\textbf{$\delta$=0.01}}  &
  \multicolumn{1}{c|}{\textbf{$\delta$=0.1}}  &
  \multicolumn{1}{c|}{\textbf{$\delta$=0.01}}  &
  \multicolumn{1}{c|}{\textbf{$\delta$ =0.1}} &
  \multicolumn{1}{c|}{\textbf{$\delta$=0.01}}  &
  \multicolumn{1}{c|}{\textbf{$\delta$=0.1}}  &
  \textbf{$\delta$=0.01} &
   &
   \\ \cline{1-11}
a &
  \multicolumn{1}{c|}{0.199} &
  \multicolumn{1}{c|}{0.622} &
  \multicolumn{1}{c|}{0.125} &
  \multicolumn{1}{c|}{0.558} &
  \multicolumn{1}{c|}{0.199} &
  \multicolumn{1}{c|}{0.663} &
  \multicolumn{1}{c|}{0.663} &
  \multicolumn{1}{c|}{0.663} &
  \multicolumn{1}{c|}{0.523} &
  0.663 &
   &
   \\ \cline{1-11}
b &
  \multicolumn{1}{c|}{0.143} &
  \multicolumn{1}{c|}{0.380} &
  \multicolumn{1}{c|}{0.131} &
  \multicolumn{1}{c|}{0.387} &
  \multicolumn{1}{c|}{0.144} &
  \multicolumn{1}{c|}{0.440} &
  \multicolumn{1}{c|}{0.586} &
  \multicolumn{1}{c|}{0.586} &
  \multicolumn{1}{c|}{0.340} &
  0.586 &
   &
   \\ \cline{1-11}
c &
  \multicolumn{1}{c|}{0.365} &
  \multicolumn{1}{c|}{0.630} &
  \multicolumn{1}{c|}{0.331} &
  \multicolumn{1}{c|}{0.675} &
  \multicolumn{1}{c|}{0.320} &
  \multicolumn{1}{c|}{0.676} &
  \multicolumn{1}{c|}{0.676} &
  \multicolumn{1}{c|}{0.676} &
  \multicolumn{1}{c|}{0.489} &
  0.676 &
   &
   \\ \cline{1-11}
d &
  \multicolumn{1}{c|}{0.236} &
  \multicolumn{1}{c|}{0.511} &
  \multicolumn{1}{c|}{0.161} &
  \multicolumn{1}{c|}{0.549} &
  \multicolumn{1}{c|}{0.161} &
  \multicolumn{1}{c|}{0.489} &
  \multicolumn{1}{c|}{0.623} &
  \multicolumn{1}{c|}{0.623} &
  \multicolumn{1}{c|}{0.436} &
  0.623 &
  \multicolumn{1}{c}{} &
  \multicolumn{1}{c}{} \\ \cline{1-11}
\textbf{e} &
  \multicolumn{1}{c|}{\textbf{0.141}} &
  \multicolumn{1}{c|}{\textbf{0.348}} &
  \multicolumn{1}{c|}{\textbf{0.102}} &
  \multicolumn{1}{c|}{0.532} &
  \multicolumn{1}{c|}{\textbf{0.080}} &
  \multicolumn{1}{c|}{\textbf{0.424}} &
  \multicolumn{1}{c|}{0.710} &
  \multicolumn{1}{c|}{0.710} &
  \multicolumn{1}{c|}{\textbf{0.321}} &
  0.641 &
   &
   \\ \cline{1-11}
f &
  \multicolumn{1}{c|}{0.199} &
  \multicolumn{1}{c|}{0.455} &
  \multicolumn{1}{c|}{0.127} &
  \multicolumn{1}{c|}{0.551} &
  \multicolumn{1}{c|}{0.125} &
  \multicolumn{1}{c|}{0.533} &
  \multicolumn{1}{c|}{0.675} &
  \multicolumn{1}{c|}{0.675} &
  \multicolumn{1}{c|}{0.328} &
  0.579 &
  \multicolumn{1}{c}{} &
  \multicolumn{1}{c}{} \\ \cline{1-11}
g &
  \multicolumn{1}{c|}{0.158} &
  \multicolumn{1}{c|}{0.373} &
  \multicolumn{1}{c|}{0.106} &
  \multicolumn{1}{c|}{0.532} &
  \multicolumn{1}{c|}{0.209} &
  \multicolumn{1}{c|}{0.532} &
  \multicolumn{1}{c|}{0.586} &
  \multicolumn{1}{c|}{\textbf{0.586}} &
  \multicolumn{1}{c|}{0.348} &
  0.586 &
   &
   \\ \cline{1-11}
h &
  \multicolumn{1}{c|}{0.330} &
  \multicolumn{1}{c|}{0.580} &
  \multicolumn{1}{c|}{0.138} &
  \multicolumn{1}{c|}{0.682} &
  \multicolumn{1}{c|}{0.093} &
  \multicolumn{1}{c|}{0.486} &
  \multicolumn{1}{c|}{0.724} &
  \multicolumn{1}{c|}{0.724} &
  \multicolumn{1}{c|}{0.486} &
  0.724 &
   &
   \\ \cline{1-11}
i &
  \multicolumn{1}{c|}{0.214} &
  \multicolumn{1}{c|}{0.408} &
  \multicolumn{1}{c|}{0.174} &
  \multicolumn{1}{c|}{0.476} &
  \multicolumn{1}{c|}{0.149} &
  \multicolumn{1}{c|}{0.442} &
  \multicolumn{1}{c|}{\textbf{0.573}} &
  \multicolumn{1}{c|}{0.573} &
  \multicolumn{1}{c|}{0.396} &
  \textbf{0.573} &
   &
   \\ \cline{1-11}
j &
  \multicolumn{1}{c|}{0.221} &
  \multicolumn{1}{c|}{0.465} &
  \multicolumn{1}{c|}{0.187} &
  \multicolumn{1}{c|}{0.596} &
  \multicolumn{1}{c|}{0.141} &
  \multicolumn{1}{c|}{0.457} &
  \multicolumn{1}{c|}{0.776} &
  \multicolumn{1}{c|}{0.776} &
  \multicolumn{1}{c|}{0.475} &
  0.682 &
   &
   \\ \cline{1-11}
k &
  \multicolumn{1}{c|}{0.243} &
  \multicolumn{1}{c|}{0.498} &
  \multicolumn{1}{c|}{0.194} &
  \multicolumn{1}{c|}{0.557} &
  \multicolumn{1}{c|}{0.146} &
  \multicolumn{1}{c|}{0.513} &
  \multicolumn{1}{c|}{0.794} &
  \multicolumn{1}{c|}{0.794} &
  \multicolumn{1}{c|}{0.467} &
  0.707 &
  \multicolumn{1}{c}{} &
  \multicolumn{1}{c}{} \\ \cline{1-11}
\end{tabular}%
\label{allmodelBIN}
\end{table*}

\textbf{Fused Classification.} For this approach, the optimal range is observed at alignment settings from \textit{d} to \textit{i}, with \textit{g} being possibly the optimal case. 
At the same time, this methodology allows for superior results in comparison to the binary classification case, which may be related to the regularization imposed by the face recognition task.
\begin{table*}[ht!]
\caption{BPCER@APCER = (0.1, 0.01) of S-MAD fused approach for various alignment settings}
\centering
\begin{tabular}{|c|cccccccccccc|}
\hline
\multirow{3}{*}{\textbf{Alignments}} &
  \multicolumn{10}{c|}{\textbf{BPCER@APCER=$\delta$}} \\ \cline{2-11} 
 &
  \multicolumn{2}{c|}{\textbf{Protocol-asml}} &
  \multicolumn{2}{c|}{\textbf{Protocol-facemorpher}} &
  \multicolumn{2}{c|}{\textbf{Protocol-opencv}} &
  \multicolumn{2}{c|}{\textbf{Protocol-stylegan}} &
  \multicolumn{2}{c|}{\textbf{Protocol-webmorph}} \\ \cline{2-11} 
 &
  \multicolumn{1}{c|}{\textbf{$\delta$=0.1}} &
  \multicolumn{1}{c|}{\textbf{$\delta$=0.01}} &
  \multicolumn{1}{c|}{\textbf{$\delta$=0.1}} &
  \multicolumn{1}{c|}{\textbf{$\delta$=0.01}} &
  \multicolumn{1}{c|}{\textbf{$\delta$=0.1}} &
  \multicolumn{1}{c|}{\textbf{$\delta$=0.01}} &
  \multicolumn{1}{c|}{\textbf{$\delta$ =0.1}} &
  \multicolumn{1}{c|}{\textbf{$\delta$=0.01}} &
  \multicolumn{1}{c|}{\textbf{$\delta$=0.1}} &
  \multicolumn{1}{c|}{\textbf{$\delta$=0.01}} \\ \hline
a &
  \multicolumn{1}{c|}{0.159} &
  \multicolumn{1}{c|}{0.689} &
  \multicolumn{1}{c|}{0.187} &
  \multicolumn{1}{c|}{0.517} &
  \multicolumn{1}{c|}{0.239} &
  \multicolumn{1}{c|}{0.599} &
  \multicolumn{1}{c|}{0.842} &
  \multicolumn{1}{c|}{0.946} &
  \multicolumn{1}{c|}{0.606} &
  \multicolumn{1}{c|}{0.885}  \\ \hline
b &
  \multicolumn{1}{c|}{0.063} &
  \multicolumn{1}{c|}{0.495} &
  \multicolumn{1}{c|}{0.072} &
  \multicolumn{1}{c|}{0.646} &
  \multicolumn{1}{c|}{0.099} &
  \multicolumn{1}{c|}{0.658} &
  \multicolumn{1}{c|}{0.671} &
  \multicolumn{1}{c|}{0.946} &
  \multicolumn{1}{c|}{0.702} &
  \multicolumn{1}{c|}{0.964}  \\ \hline
c &
  \multicolumn{1}{c|}{0.125} &
  \multicolumn{1}{c|}{0.467} &
  \multicolumn{1}{c|}{0.215} &
  \multicolumn{1}{c|}{0.588} &
  \multicolumn{1}{c|}{0.240} &
  \multicolumn{1}{c|}{0.566} &
  \multicolumn{1}{c|}{0.694} &
  \multicolumn{1}{c|}{0.884} &
  \multicolumn{1}{c|}{0.541} &
  \multicolumn{1}{c|}{0.859}  \\ \hline
d &
  \multicolumn{1}{c|}{0.040} &
  \multicolumn{1}{c|}{0.374} &
  \multicolumn{1}{c|}{0.102} &
  \multicolumn{1}{c|}{0.558} &
  \multicolumn{1}{c|}{0.103} &
  \multicolumn{1}{c|}{0.568} &
  \multicolumn{1}{c|}{0.574} &
  \multicolumn{1}{c|}{0.835} &
  \multicolumn{1}{c|}{0.305} &
  \multicolumn{1}{c|}{0.781} \\ \hline
e &
  \multicolumn{1}{c|}{0.162} &
  \multicolumn{1}{c|}{0.580} &
  \multicolumn{1}{c|}{0.149} &
  \multicolumn{1}{c|}{0.582} &
  \multicolumn{1}{c|}{0.177} &
  \multicolumn{1}{c|}{0.602} &
  \multicolumn{1}{c|}{0.566} &
  \multicolumn{1}{c|}{0.767} &
  \multicolumn{1}{c|}{0.605} &
  \multicolumn{1}{c|}{0.870} \\ \hline
f &
  \multicolumn{1}{c|}{0.184} &
  \multicolumn{1}{c|}{0.530} &
  \multicolumn{1}{c|}{0.180} &
  \multicolumn{1}{c|}{0.488} &
  \multicolumn{1}{c|}{0.175} &
  \multicolumn{1}{c|}{0.451} &
  \multicolumn{1}{c|}{0.582} &
  \multicolumn{1}{c|}{0.788} &
  \multicolumn{1}{c|}{0.517} &
  \multicolumn{1}{c|}{0.785} \\ \hline
\textbf{g} &
  \multicolumn{1}{c|}{\textbf{0.034}} &
  \multicolumn{1}{c|}{\textbf{0.233}} &
  \multicolumn{1}{c|}{\textbf{0.025}} &
  \multicolumn{1}{c|}{0.701} &
  \multicolumn{1}{c|}{\textbf{0.037}} &
  \multicolumn{1}{c|}{0.701} &
  \multicolumn{1}{c|}{0.487} &
  \multicolumn{1}{c|}{0.875} &
  \multicolumn{1}{c|}{\textbf{0.216}} &
  \multicolumn{1}{c|}{0.788}\\ \hline
h &
  \multicolumn{1}{c|}{0.168} &
  \multicolumn{1}{c|}{0.642} &
  \multicolumn{1}{c|}{0.168} &
  \multicolumn{1}{c|}{0.535} &
  \multicolumn{1}{c|}{0.165} &
  \multicolumn{1}{c|}{0.599} &
  \multicolumn{1}{c|}{0.536} &
  \multicolumn{1}{c|}{0.850} &
  \multicolumn{1}{c|}{0.542} &
  \multicolumn{1}{c|}{0.854}  \\ \hline
i &
  \multicolumn{1}{c|}{0.046} &
  \multicolumn{1}{c|}{0.255} &
  \multicolumn{1}{c|}{0.036} &
  \multicolumn{1}{c|}{\textbf{0.365}} &
  \multicolumn{1}{c|}{0.044} &
  \multicolumn{1}{c|}{\textbf{0.390}} &
  \multicolumn{1}{c|}{\textbf{0.305}} &
  \multicolumn{1}{c|}{\textbf{0.583}} &
  \multicolumn{1}{c|}{0.246} &
  \multicolumn{1}{c|}{\textbf{0.554}}  \\ \hline
j &
  \multicolumn{1}{c|}{0.287} &
  \multicolumn{1}{c|}{0.630} &
  \multicolumn{1}{c|}{0.268} &
  \multicolumn{1}{c|}{0.585} &
  \multicolumn{1}{c|}{0.262} &
  \multicolumn{1}{c|}{0.564} &
  \multicolumn{1}{c|}{0.844} &
  \multicolumn{1}{c|}{0.959} &
  \multicolumn{1}{c|}{0.697} &
  \multicolumn{1}{c|}{0.907} \\ \hline
k &
  \multicolumn{1}{c|}{0.193} &
  \multicolumn{1}{c|}{0.652} &
  \multicolumn{1}{c|}{0.253} &
  \multicolumn{1}{c|}{0.745} &
  \multicolumn{1}{c|}{0.262} &
  \multicolumn{1}{c|}{0.792} &
  \multicolumn{1}{c|}{0.825} &
  \multicolumn{1}{c|}{0.953} &
  \multicolumn{1}{c|}{0.674} &
  \multicolumn{1}{c|}{0.915} \\ \hline
\end{tabular}%
    \label{tabre}
\end{table*}
\begin{figure}[ht!]
\centering
    \includegraphics[width=0.98\linewidth]{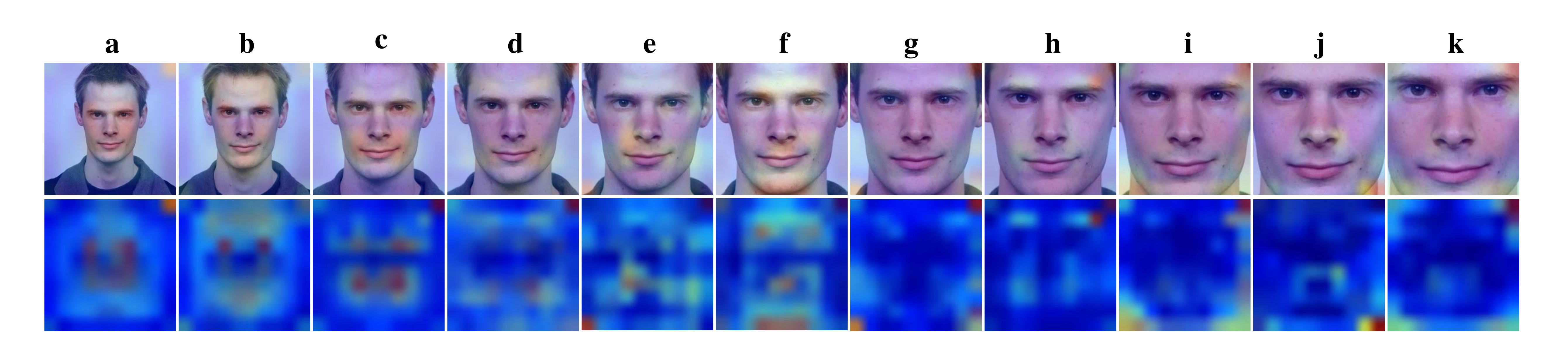}
    \caption{\label{grad1} Grad-CAM morph heatmaps for the S-MAD fused approach under different alignment conditions (Recall that \textit{bona fide} sets are equal across all the protocols).}
\end{figure}
Based on the heatmaps, the detection is mainly focused on the face region and, in many cases, on the regions of intersection between the foreground and background (see Fig. \ref{grad1}). 

\textbf{NIST FRVT MORPH Results.} We compare the results of our best model (\textbf{visteamicao-000}) for fused case with several state-of-the-art (SOTA) MAD approaches, tested on the FRVT NIST MORPH Benchmark \cite{frvtmorphnist}. Each dataset from the benchmark has images generated through different protocols, with distinctions made in tiers such as Tier 2 - Automated Morph Analysis and Tier 3 - High-Quality Morph Analysis.

\begin{table*}[ht!]
\centering
 \caption{Comparison with the SOTA S-MAD approaches using APCER@BPCER = (0.1, 0.01).}
 \label{tab:nist}
\begin{tabular}{|cl|cc|clc|clc|}
\hline
\multicolumn{2}{|c|}{\multirow{2}{*}{\textbf{Algorithm}}} &
  \multicolumn{2}{c|}{\textbf{Visa-Border (Tier 2)}} &
  \multicolumn{3}{c|}{\textbf{Twente (Tier 2)}} &
  \multicolumn{3}{c|}{\textbf{Manual (Tier 3)}} \\ \cline{3-10} 
\multicolumn{2}{|c|}{} &
  \multicolumn{1}{c|}{\textbf{$\delta$=0.1}} &
  \textbf{$\delta$=0.01} &
  \multicolumn{2}{c|}{\textbf{$\delta$=0.1}} &
  \textbf{$\delta$=0.01} &
  \multicolumn{2}{c|}{\textbf{$\delta$=0.1}} &
  \textbf{$\delta$=0.01} \\ \hline
\multicolumn{2}{|c|}{Our} & \multicolumn{1}{c|}{0.089} & \textbf{0.291} & \multicolumn{2}{c|}{0.032} & 0.128 & \multicolumn{2}{c|}{0.802} & 0.975 \\ \hline
\multicolumn{2}{|c|}{Aghdaie et al. \cite{aghdaie2021attention}}            & \multicolumn{1}{c|}{0.037} & 0.542 & \multicolumn{2}{c|}{0.002} & \textbf{0.060} & \multicolumn{2}{c|}{0.879} & 0.975 \\ \hline
\multicolumn{2}{|c|}{Medvedev et al. \cite{MorDeephy}}              & \multicolumn{1}{c|}{0.232} & 0.555 & \multicolumn{2}{c|}{0.174} & 0.493 & \multicolumn{2}{c|}{0.641} & \textbf{0.926} \\ \hline
\multicolumn{2}{|c|}{Ferrara et al. \cite{unibo}}                & \multicolumn{1}{c|}{0.477} & 0.999 & \multicolumn{2}{c|}{0.002} & 0.183 & \multicolumn{2}{c|}{0.938} & 0.985 \\ \hline
\multicolumn{2}{|c|}{Ramachandra et al.\cite{ramachandra2019morphing}}              & \multicolumn{1}{c|}{0.375} & 0.990 & \multicolumn{2}{c|}{0.304} & 0.998 & \multicolumn{2}{c|}{0.938} & 0.985 \\ \hline
\end{tabular}%
\end{table*}

\begin{figure*}[ht!]
 \centering
 \captionsetup{justification=centering}
 \includegraphics[width=0.86\linewidth]{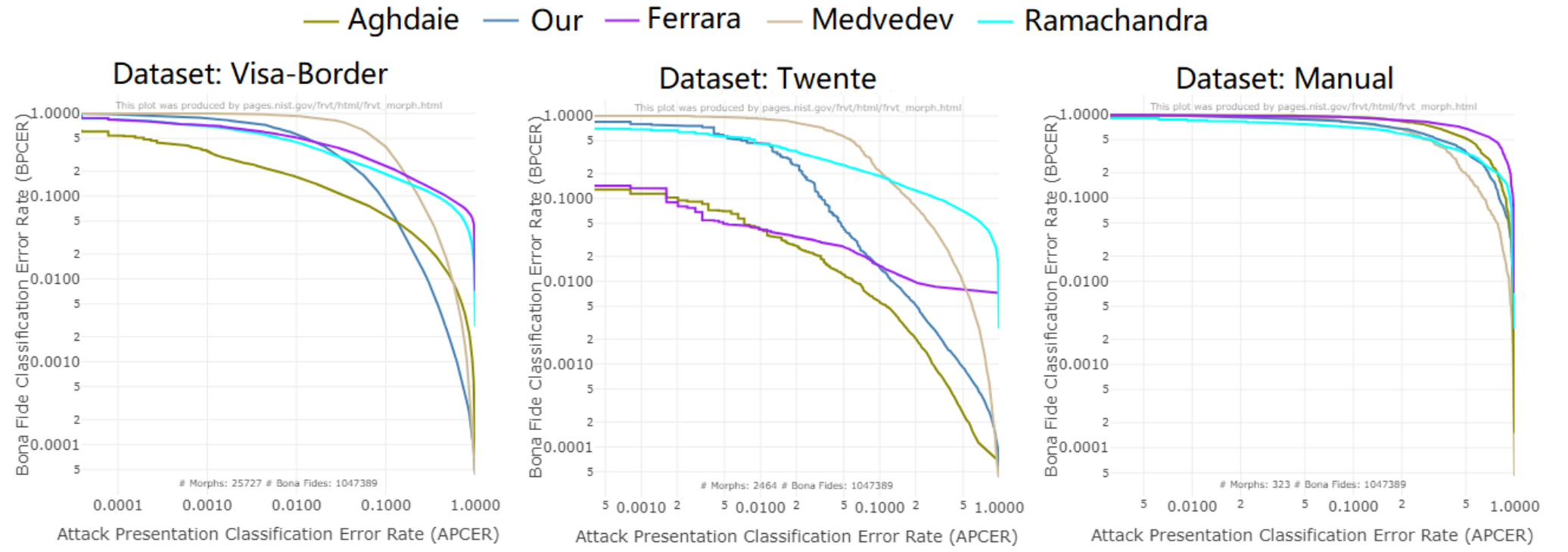}
  \caption{Detection Error Trade-off curves for different SOTA approaches in different datasets (Visa-Border, Twente and Manual dataset).} 
     \label{fig:printscaneds}
 \end{figure*}

Regarding the Visa-Border dataset, our approach outperforms all other SOTA approaches, with a \textit{morph miss rate} of 0.29 at a \textit{false detection} rate of 0.01.
In the Twente dataset, when comparing with other approaches, the results demonstrate a highly favorable outcome as well, with a \textit{morph error rate} of 0.128 at a \textit{false detection rate} of 0.01 (See table \ref{tab:nist}).
Although not represented in the table, comparable results were achieved for other datasets, such as the UNIBO Automatic Morphed
Face Generation Tool v1.0 and even MIPGAN-II with less dominant but still competitive performances.

It is important to take into consideration the influence of the dataset used, and this Tier 2 typology is generally less challenging. 
When faced with more realistic datasets (Manual dataset), it becomes apparent that overall SOTA approaches show poor generalization across various unseen morphing techniques. Even so, our model results achieved competitive results when compared to those approaches.
\vspace{0cm}
\section{Conclusions}
\vspace{0cm}

In this work, we aim to identify the context properties that are most effective for S-MAD.
The extensive experiments allowed us to determine the alignment range where S-MAD is more effective. Moreover, in this range, there seems to be a certain correspondence between both fused and binary approaches, which translates into a similar area of face occupancy in the image. 
Despite that, our results also show that face is the most dominant activation region across all the alignment settings, and the impact of context on face morphing detection is limited. 
Our method achieved state-of-the-art comparable performances on some of the NIST FRVT MORPH benchmark protocols.
Our future work will be directed toward investigating similar properties in the differential scenario.

\section{Acknowledgements}
The authors would like to thank the Portuguese Mint and Official Printing Office (INCM) and the Institute of Systems and Robotics - the University of Coimbra for the support of the project FACING2. This work has been supported by Fundação para a Ciência e a Tecnologia (FCT) under the project UIDB/00048/2020.

\bibliography{bibliography.bib}

\bibliographystyle{IEEEtran}

\end{document}